\documentclass[sigconf]{acmart}

\AtBeginDocument{%
  }

\usepackage{tikz}
\usetikzlibrary{shapes.geometric, arrows.meta, positioning, fit, backgrounds}
\usepackage{amsmath}

\usepackage{amssymb}
\usepackage{multirow} 

\setcopyright{acmlicensed}
\copyrightyear{2026}
\acmYear{2026}
\acmDOI{XXXXXXX.XXXXXXX}

\acmConference[ICAIL '26]{Twenty-First International Conference on Artificial Intelligence and Law}{June 08--12, 2026}{Singapore, Singapore}

\begin{document}

\title{Learning When Not to Decide: A Framework for Overcoming Factual Presumptuousness in AI Adjudication}

\author{Mohamed Afane}
\email{afane@law.stanford.edu}
\affiliation{%
  \institution{Stanford University}
  \city{Stanford}
  \state{CA}
  \country{USA}}

\author{Emily Robitschek}
\email{erobitsc@law.stanford.edu}
\affiliation{%
  \institution{Stanford University}
  \city{Stanford}
  \state{CA}
  \country{USA}}

\author{Derek Ouyang}
\email{douyang1@law.stanford.edu}
\affiliation{%
  \institution{Stanford University}
  \city{Stanford}
  \state{CA}
  \country{USA}}

\author{Daniel E. Ho}
\email{dho@law.stanford.edu}
\affiliation{%
  \institution{Stanford University}
  \city{Stanford}
  \state{CA}
  \country{USA}}

\renewcommand{\shortauthors}{Afane, Robitschek, Ouyang, and Ho}
\begin{abstract}
A well-known limitation of AI systems is \emph{presumptuousness}: the tendency of AI systems to provide confident answers when information may be lacking. This challenge is particularly acute in legal applications, where a core task for attorneys, judges, and administrators is to determine whether evidence is sufficient to reach a conclusion. We study this problem in the important setting of unemployment insurance adjudication, which has seen rapid integration of AI systems and where the question of additional fact-finding poses the most significant bottleneck for a system that affects millions of applicants annually. First, through a collaboration with the Colorado Department of Labor and Employment, we secure rare access to official training materials and guidance to design a novel benchmark that systematically varies in information completeness. Second, we evaluate four leading AI platforms and show that standard RAG-based approaches achieve an average of only 15\% accuracy when information is insufficient. Third, advanced prompting methods improve accuracy on inconclusive cases but over-correct, withholding decisions even on clear cases. Fourth, we introduce a structured framework requiring explicit identification of missing information before any determination (SPEC, \underline{S}tructured \underline{P}rompting for \underline{E}vidence \underline{C}hecklists). SPEC achieves 89\% overall accuracy, while appropriately deferring when evidence is insufficient---demonstrating that presumptuousness in legal AI is systematic but addressable, and that doing so is a necessary step towards systems that reliably support, rather than supplant, human judgment wherever decisions must await sufficient evidence.
\end{abstract}




\maketitle

\section{Introduction}
\label{sec:intro}

Legal reasoning has always required navigating uncertainty. Statutes employ vague standards---``reasonable efforts,'' ``good cause,'' ``willful misconduct''---that demand interpretation against specific facts. Evidence is incomplete; witnesses contradict each other; circumstances resist clean categorization. For human adjudicators, this is familiar terrain. For artificial intelligence (AI), it has proven treacherous. 

The earliest legal expert systems, developed in the 1980s and 1990s, encoded statutory rules as deterministic decision trees. They worked well when law reduced to bright-line tests---twenty weeks of employment, a specific dollar threshold---but collapsed when confronting the uncertainty  endemic to legal reasoning. Governing law proved too complex to represent directly as symbolic logic; as Kevin Ashley documented, they ``did not possess the capacity to represent uncertainty or engage in non-monotonic reasoning about changing facts'' \cite{ashleyArtificialIntelligenceLegal2017}. Modern transformer-based architectures have not resolved this limitation. Despite remarkable linguistic capabilities, large language models (LLM) generate fluent, assured outputs regardless of whether the underlying evidence warrants such confidence \cite{yeBenchmarkingLLMsUncertainty2024}. The systems that once failed because they could not handle uncertainty now fail because they cannot recognize insufficiency. 

This presumptuousness is a general problem for AI deployed in reasoning tasks, but it is particularly acute in law---and nowhere more so than in administrative adjudication, where AI deployment is accelerating, and errors directly harm individuals  \cite{engstromAlgorithmicAccountabilityAdministrative2020a}. Unemployment insurance (UI) offers a compelling domain to study this problem. The governing law is formidably complex: scattered statutory provisions, vague eligibility standards, and jurisdiction-specific exceptions that resist systematic codification \cite{pahlkaRecodingAmericaWhy2023, youngUsingArtificialIntelligence2022}. Developing sound adjudicative judgment can take years. When a researcher asked a California UI claims processor why he kept deferring questions to senior colleagues, he replied: ``I've only been here 17 years. The folks who really know how UI works have been here for 25 years or more'' \cite{pahlkaRecodingAmericaWhy2023}. Agencies maintain extensive training programs to cultivate this judgment in new adjudicators---and increasingly look to AI to help.

The pressure to automate is intense. During the pandemic, unemployment systems nationwide collapsed under unprecedented claim volumes, not primarily because they lacked processing capacity, but because fact-finding became the bottleneck. Determining eligibility requires gathering and weighing evidence: why did the claimant leave their job? Did the employer provide adequate documentation? Is the claimant's account consistent with the record? When systems could not perform this evidentiary work adequately, they failed \cite{pahlkaRecodingAmericaWhy2023}. Agencies find themselves caught between competing demands: rely on merit staff and fail to meet timeliness requirements, or automate and risk running afoul of rules designed to ensure human judgment \cite{hoEvaluationDueProcess}. The Colorado Department of Labor and Employment (CDLE) has piloted AI tools designed to help adjudicators ``do the work more accurately and efficiently'' rather than replace human judgment \cite{almsLaborDepartmentExperiments2024}. New Hampshire is testing an adjudication assistant that collects more complete information at the start of claims \cite{StateNewHampshire2023}. The question is not whether AI will be integrated into benefits administration---it already is---but whether these systems can recognize the limits of their own knowledge.

The consequences of getting this wrong are severe. In 2015, Michigan's automated UI system falsely accused over 20,000 residents of fraud, issuing confident determinations without adequate evidentiary support. The system's designers had not equipped it to recognize when information was insufficient; it simply decided. Many residents had wages garnished or tax refunds seized before the errors came to light. At least one person took her own life after being assessed \$50,000 in penalties \cite{feltonMichiganUnemploymentAgency2016}. The pattern extends to eligibility determinations: AI systems that deny without sufficient evidence strand workers who may qualify; systems that approve without sufficient evidence create overpayments that agencies later claw back. Either way, the harm flows from deciding before the facts warrant it.

A common approach to addressing these concerns is to have a human in the loop. But this framing obscures critical design choices. Two well-known failure modes in human--AI interaction are automation bias, where humans over-rely on algorithmic suggestions, and algorithm aversion, where humans dismiss useful recommendations \cite{alon-barkatHumanAIInteractions2023}. Humans can identify erroneous algorithmic outputs and override them, but only when systems provide clear signals about uncertainty rather than overconfident predictions that humans must second-guess \cite{de-arteagaCaseHumansintheLoopDecisions2020}. A system that explicitly identifies missing information before reaching a determination provides exactly such a signal.

Most benchmarks for evaluating AI systems, including those in legal domains, sidestep this question entirely. They assess performance when all relevant information is provided, testing whether models reach correct conclusions, not whether they can recognize when conclusions are unwarranted \cite{guhaLegalBenchCollaborativelyBuilt2023a, hendrycksCUADExpertAnnotatedNLP, holzenbergerDatasetStatutoryReasoning2020, zhengWhenDoesPretraining2021a, koreedaContractNLIDatasetDocumentlevel2021}. This paper addresses that gap. Through a collaboration with CDLE, we obtain rare access to official training materials---the actual guidance used to teach human adjudicators when they have enough information to decide. We use these materials, alongside governing statutes provided through retrieval-augmented generation (RAG), to construct eligibility scenarios that vary systematically in information completeness: cases that are clearly eligible, clearly ineligible, and cases where a competent adjudicator should recognize that determination must await further fact-finding.

We contribute: \textit{(1)} The first legal adjudication dataset that precisely controls the presence or absence of legally operative facts, developed through collaboration with CDLE that provided deep immersion into the domain, including thorough review of internal training materials and comprehensive understanding of how information gaps manifest across diverse cases; \textit{(2)} An evaluation showing that four leading AI platforms, given identical statutory and official adjudication guidance through RAG, achieve on average 15\% accuracy on cases lacking sufficient information, delivering presumptuous yes/no eligibility decisions when they should defer; \textit{(3)} Evidence that advanced prompting methods and detailed adjudication instructions, even with explicit guidance to withhold decisions when information is missing, over-correct and fail to decide clear cases; and \textit{(4)} SPEC (\underline{S}tructured \underline{P}rompting for \underline{E}vidence \underline{C}hecklists), a multi-stage framework that eliminates baseline presumptuousness and resolves the determination--deferral tradeoff, achieving 89\% overall accuracy with appropriate deferral when evidence is insufficient.

\section{Unemployment Insurance}
\label{sec:ui_adjudication}

Unemployment insurance is an emblematic case of legal reasoning amid incomplete information.
UI provides temporary income support to workers who lose employment through no fault of their own. Eligibility determinations require applying statutory criteria to case-specific facts. For example, under C.R.S. § 8-73-108, a claimant discharged for workplace misconduct may be disqualified from benefits, but only if the conduct meets statutory definitions requiring assessment of severity, intent, and employer policies. 

Adjudicators work from information submitted by claimants and employers, supplemented by interviews and documentation requests. This information is routinely incomplete: claimants may not know what facts are legally relevant; employers may provide conclusory characterizations without supporting detail; accounts may conflict on outcome-determinative points. Determining when sufficient information exists to decide, versus when additional fact-finding is required, is central to adjudicative practice.

State agencies maintain detailed adjudication guides that operationalize statutory requirements, specifying what facts must be established for each eligibility criterion, how to weigh conflicting evidence, and when to request additional information. In a unique collaboration, we are able to access CDLE's guide, which covers voluntary quit, discharge for misconduct, availability for work, and other criteria under Colorado Revised Statutes. 

\section{Related Works}
\label{sec:background}

Our work pertains to five bodies of research. 

\textbf{1. Retrieval-augmented generation (RAG) for Legal Tasks.} RAG addresses a fundamental limitation of language models: their knowledge is fixed at training time \cite{lewisRetrievalAugmentedGenerationKnowledgeIntensive2020a} and may not include the specific legal provisions relevant to a given task. RAG systems retrieve relevant documents---statutes, regulations, adjudication guides---and provide them to LLMs as context alongside each query, grounding model outputs in authoritative sources \cite{gaoRetrievalAugmentedGenerationLarge2023, radityaRetrievalaugmentedGenerationRAG2025}. 

For legal adjudication, RAG offers a natural fit. The system can retrieve exactly those statutory provisions and interpretive guidance that govern the case at hand, rather than relying on whatever legal knowledge the model absorbed during pretraining. Recent work has explored sophisticated retrieval strategies, including case-based reasoning approaches that use similarity between legal questions to improve context selection \cite{wiratungaCBRRAGCaseBasedReasoning2024}. In our evaluation, all systems receive identical Colorado UI materials through RAG: the relevant Colorado Revised Statutes, administrative regulations, and official adjudication guidance used by CDLE.

However, retrieval alone does not overcome presumptuousness \cite{mageshHallucinationFreeAssessingReliability2024}. Providing a model with the legal standard for ``willful misconduct'' does not ensure it will recognize when the facts provided are insufficient to apply that standard. The model may retrieve the correct legal rule and still issue a determination based on incomplete evidence. Our evaluation tests the rate at which this occurs and whether structured prompting can prevent it. 

\textbf{2. Legal AI Benchmarks.} Existing legal AI benchmarks designed specifically for legal reasoning---LegalBench's statutory and contract  reasoning tasks \cite{guhaLegalBenchCollaborativelyBuilt2023a}, CUAD's contract clause extraction \cite{hendrycksCUADExpertAnnotatedNLP}, SARA's tax law  entailment tasks \cite{holzenbergerDatasetStatutoryReasoning2020}, CaseHOLD's legal holding identification \cite{zhengWhenDoesPretraining2021a}, and  ContractNLI's contract inference evaluation \cite{koreedaContractNLIDatasetDocumentlevel2021}---all share a structural assumption: that sufficient information exists to determine the correct answer. None systematically test whether models can recognize when information is \textit{insufficient}: when the appropriate response is ``more facts are needed,'' rather than a definitive conclusion. This gap mirrors a broader limitation in LLM evaluation, where benchmarks predominantly assess passive reasoning with complete information rather than active reasoning requiring information gathering \cite{zhouPassiveActiveReasoning2025}. Recent work has begun examining related limitations. He et al. address \textit{interpretive ambiguity}---the challenge arising when legal principles can be read multiple ways---proposing computational frameworks that mirror legal mechanisms for constraining interpretation \cite{heStatutoryConstructionInterpretation2025a}. Our work addresses a complementary problem: not ambiguity in what the rule means, but insufficiency in the facts available to apply it. Both require AI systems to overcome presumptuousness, but operate at different points in the reasoning process.

\begin{figure}[t]
    \centering
    \includegraphics[width=0.5\textwidth]{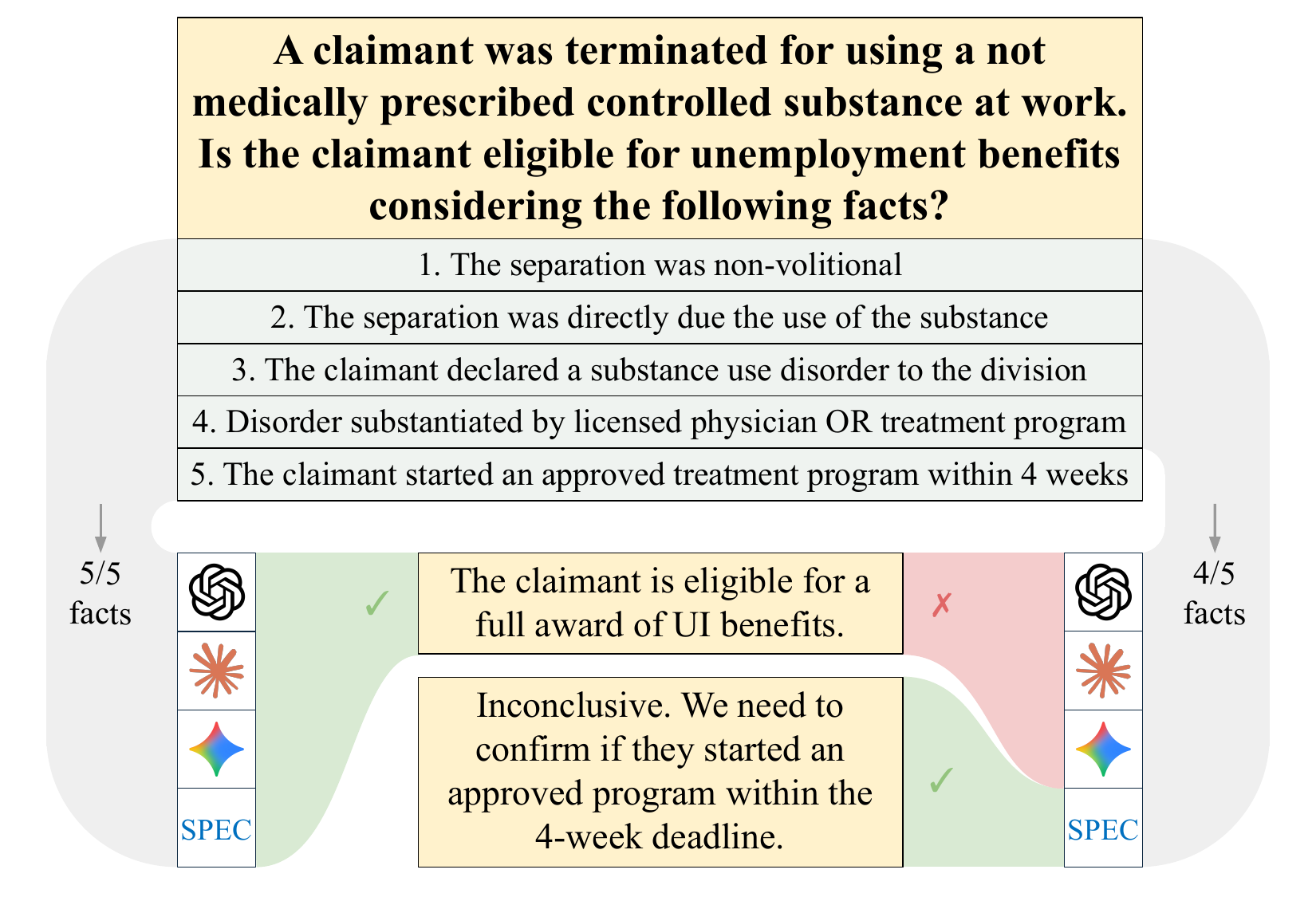}
    \caption{Comparison of model outputs on a termination case involving non-prescribed controlled substance use at work. When all five statutory requirements are provided (left), models reach the correct eligibility determination. When one requirement is withheld (right), standard LLMs still issue presumptuous eligibility decisions, while SPEC identifies the missing information and outputs Inconclusive.}
    \label{fig:substance-use-case}
\end{figure}

\textbf{3. LLM Information-Seeking Under Uncertainty.} A growing body of work examines whether language models can recognize information gaps in other domains. QuestBench evaluates whether models can identify missing variables in underspecified logic and math problems; even when models solve fully-specified versions, they achieve only 40--50\% accuracy at selecting appropriate clarification questions \cite{liQuestBenchCanLLMs2025}. AR-Bench tests active reasoning through iterative questioning in detective scenarios and puzzles, finding that models frequently fail to acquire needed information even with sophisticated search strategies \cite{zhouPassiveActiveReasoning2025}. MediQ simulates clinical interactions requiring diagnostic follow-up questions; directly prompting models to ask questions actually \textit{degrades} performance by 11\% compared to immediate diagnosis \cite{liMediQQuestionAskingLLMs2024}. 

These parallel efforts reveal a common problem: current LLMs struggle to recognize when they lack sufficient information to answer confidently. Wang et al. find a related pattern in comparing AI agents to human workers: agents take ``an overwhelmingly programmatic approach'' across work domains, often producing inferior work while masking deficiencies through confident execution \cite{wangHowAIAgents2025}. The machine learning literature formalizes this challenge as \textit{abstention}: knowing when not to answer \cite{wen_know_2025}. AbstentionBench finds that reasoning fine-tuning hurts abstention and model scale provides no improvement \cite{kirichenko_abstentionbench_2025}; medical QA work reaches similar conclusions \cite{machcha_knowing_2026}. The selective prediction literature formalizes the underlying tradeoff: improving accuracy on answered cases reduces coverage of decidable ones \cite{geifman_selective_2017, geifman_selectivenet_2019}, with subsequent work addressing over-abstention specifically \cite{srinivasan_selective_2024}. 

But in legal adjudication, abstention is not merely a statistical tradeoff to optimize---it is a procedural requirement. When evidence is insufficient, the appropriate response is additional fact-finding, not a confident guess. And effective human--AI collaboration depends on this distinction: when systems provide confident predictions, humans struggle to identify when override is warranted \cite{de-arteagaCaseHumansintheLoopDecisions2020}. Explicit gap identification provides exactly the signal that assertive predictions obscure. Our work extends this investigation to the legal domain, where deferral carries concrete procedural meaning, authoritative texts define what information is needed, and serious  consequences can occur when systems issue premature determinations without adequate support.

\textbf{4. Prompting Techniques and Model Calibration.} Prompting is commonly used to improve reasoning. Chain-of-Thought (CoT) decomposes complex problems into intermediate steps \cite{weiChainofthoughtPromptingElicits2022}. Tree-of-Thoughts (ToT) explores multiple reasoning branches  \cite{yaoTreeThoughtsDeliberate2023a}, extended by Graph-of-Thought to model reasoning as arbitrary graphs \cite{bestaGraphThoughtsSolving2024}. Self-Consistency uses consensus across solution paths as an uncertainty signal \cite{wangSelfConsistencyImprovesChain2023}. Reflexion enables iterative self-critique and refinement \cite{shinnReflexionLanguageAgents2023}. Multi-Agent Debate uses multiple model instances to surface errors through deliberation \cite{duImprovingFactualityReasoning2024}. 
These techniques improve reasoning \textit{quality}, but are not specifically designed to recognize \textit{information gaps}---to know when to defer rather than when to reason more carefully.

Research on model calibration addresses the related problem of aligning confidence with accuracy. Overconfidence is persistent: models generate confident predictions even when uncertainty is high \cite{yeBenchmarkingLLMsUncertainty2024}. Technical approaches including temperature scaling and learned calibration show promise \cite{desaiCalibrationPretrainedTransformers2020, shenThermometerUniversalCalibration2024}. But calibration typically asks: \textit{given the available information, how sure should the model be of its answer?} Our problem is upstream: \textit{is there enough information to answer at all?} The first relates to overconfidence within a decision; the second whether to decide in the first place, with implications for subsequent information gathering.

\textbf{5. From Symbolic Rules to Structured Guidance.} Early legal expert systems represented statutory requirements as decision trees with discrete nodes for each legally operative fact \cite{zatarainBookReviewArtificial2018}. When a required fact was unavailable, the system deferred rather than guessed. Although presumptuousness would seem to be well-addressed in this way, symbolic approaches never scaled. Translating statutes into formal logic required painstaking manual effort by legal experts and knowledge engineers working provision by provision, and the resulting systems remained brittle, unable to adapt when statutes changed or cases fell outside anticipated patterns \cite{ashleyArtificialIntelligenceLegal2017}.

Neural methods appear to offer a solution to this bottleneck. Language models appear capable of applying authoritative guidance to novel fact patterns without requiring manual formalization of every statutory provision. Even when the legal documents themselves are often dense, inconsistently organized, and difficult to operationalize directly, RAG methods help to focus LLM reasoning on the documents that contain the governing requirements, and that human adjudicators would reference.

\begin{figure}[t]
    \centering
    \includegraphics[width=0.5\textwidth]{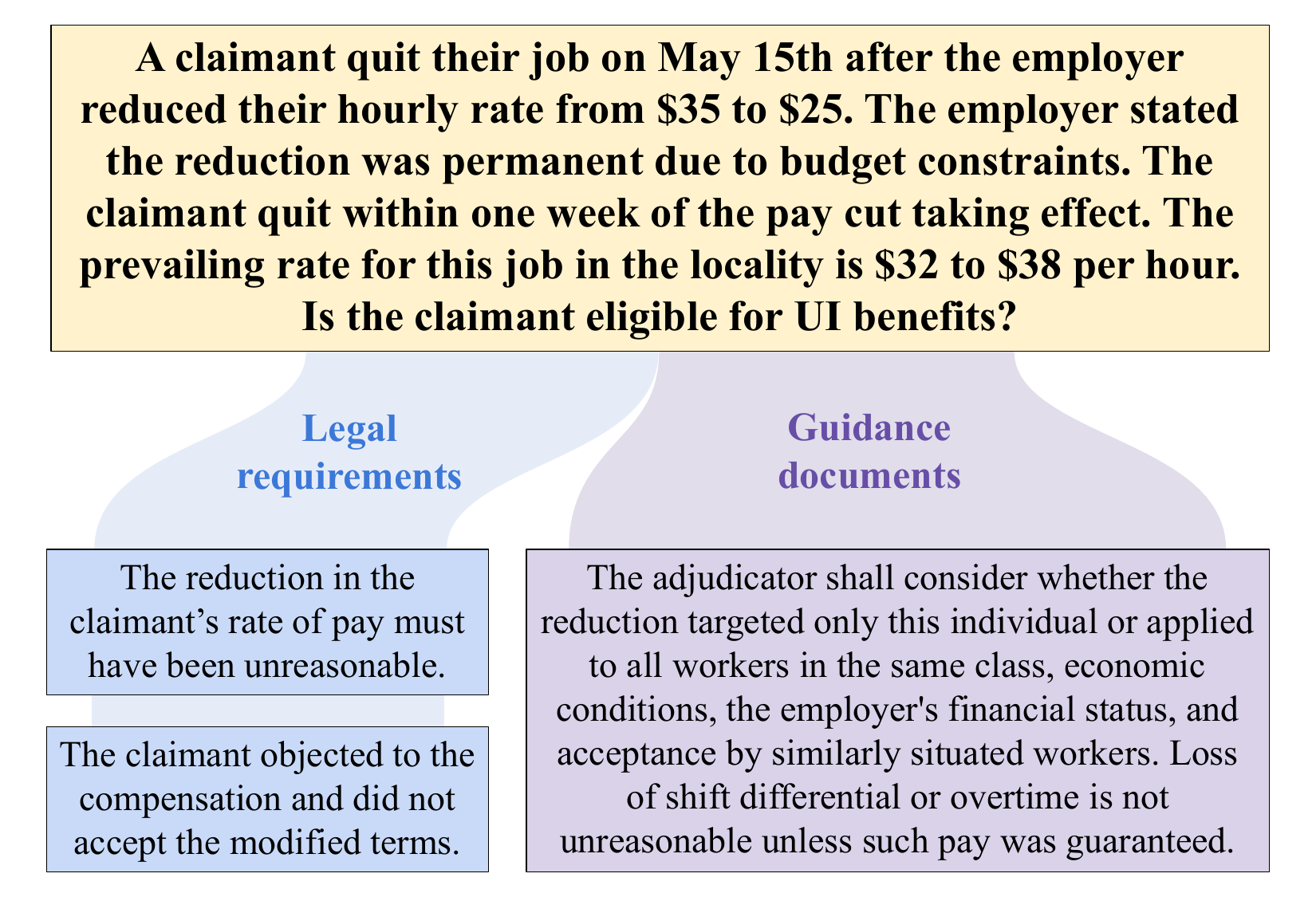}
    \caption{A voluntary quit case involving pay reduction requires contextual evaluation of two statutory requirements: whether the reduction was ``unreasonable'' and whether the claimant objected. The determination must consider multiple factors including whether the reduction targeted the individual or all workers, economic conditions, employer financial status, and acceptance by similarly situated workers.}
    \label{fig:pay-reduction-case}
\end{figure}

Yet access to the right materials is not sufficient. Figure~\ref{fig:substance-use-case} illustrates a case with discrete statutory requirements that can be verified as present or absent; when one requirement is withheld, standard language models issue presumptuous eligibility determinations despite the gap. Our work explores whether structured prompting can recover the key virtue of symbolic systems: explicit gap detection. At the same time, Figure~\ref{fig:pay-reduction-case} underscores the broader, more fundamental challenge: discrete factual requirements exist alongside qualitative standards requiring much more subjective, contextual judgment (\textit{i.e.}, whether a pay reduction was ``unreasonable''), often contained only in the state agencies' internal adjudication guides. 

SPEC offers a path forward through the seeming incompatibility of symbolic and neural systems, by recognizing that these guides approximate soft symbolic rules: not rigid decision trees, but lists identifying legally operative facts for each separation type. Neural methods can perform the contextual reasoning necessary for individual judgments---assessing whether a claimant's job search constitutes ``reasonable effort'' by weighing labor market conditions, qualifications, and search strategies---while an overarching prompting structure maintains focus on gap detection across a set of governing legal requirements and guidance standards. Thus, across different discrete stages within a legal reasoning task, our approach leverages the potential for complementarity between neural flexibility (\textit{i.e.}, distilling requirements from open-ended text, navigating qualitative judgment for soft guidance) and symbolic structure (\textit{i.e.}, maintaining a functional checklist, flagging gaps).


\section{Methodology}

SPEC is a multi-agent framework that guards against presumptuousness by making information sufficiency an explicit part of the reasoning process. The core principle is straightforward: legal determinations should only proceed when sufficient information exists to support them. When critical facts are missing, the system identifies what is needed rather than rendering an unreliable judgment.

\subsection{Problem Formulation}

Consider an adjudication task where a query $q$ describes a claimant's situation, and $C$ a corpus of statutory provisions, administrative guides, and cited case law containing the legal criteria for eligibility. The system extracts facts $F$ from $q$ and derives requirements $R$ from $C$ that must be satisfied for a determination.

For each requirement $r \in R$, the framework evaluates whether the facts $F$ adequately support $r$. 
The information gap $G$ consists of requirements lacking adequate support:
\begin{equation}
G = \{r \in R : r \text{ is not satisfied by facts $F$ in } q\}
\end{equation}
A system is properly calibrated for adjudication if it produces a determination only when no gaps exist, and otherwise outputs inconclusive along with identification of what information is missing. Standard prompting fails this criterion because language models generate determinations directly without explicitly evaluating whether each requirement is satisfied by the available facts.

\subsection{System Overview}

\begin{figure*}[t]
\centering
    \includegraphics[width=0.9\textwidth]{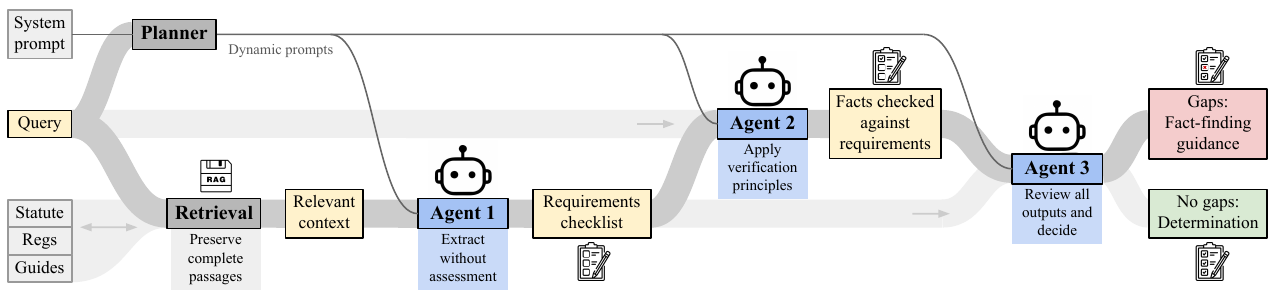}
\caption{SPEC pipeline. A RAG system extracts statutory provisions, case law, and considerations as complete passages. Dynamic prompts direct three agents: Agent 1 builds a requirements checklist, Agent 2 verifies facts against requirements, and Agent 3 provides supervisory review with authority to overturn assessments. If no critical gaps exist, a determination is made; otherwise, specific fact-finding is recommended.}
\label{fig:SPEC}
\end{figure*}

SPEC consists of three main components: requirement retrieval, multi-agent analysis, and determination. Figure~\ref{fig:SPEC} illustrates the architecture.

The retrieval component extracts all relevant passages from adjudication guides related to the query. The system retrieves complete legal citations, statutory provisions, considerations, and case law as unified blocks from $C$. Multi-part requirements are preserved together as complete passages rather than fragmented. This component performs extraction only---no conclusions or determinations are permitted. The extraction preserves exact wording to maintain legal precision and retrieves examples from adjudication guides that inform subsequent analysis.

\subsection{Multi-Agent Analysis and Fact Verification}

The core of SPEC is a three-agent system that maps requirements to facts through specialized analytical instructions. All agents must reference the exact question text without rephrasing, and treat stated facts as established without demanding external verification. 

\underline{Agent 1 (Requirements Checklist)} extracts a structured checklist from the guides with three categories: required elements (core statutory requirements with statute numbers), considerations (all considerations from the guide, noting that not all apply to every case), and case law requirements (case name and principle established). This agent performs extraction only with no assessment. Its role is to derive the requirements $R$ from $C$ that must be satisfied for a determination. The checklist provides the foundation for subsequent fact verification and supervisory review by explicitly enumerating what information must be present. 

\underline{Agent 2 (Fact Verification)} compares each checklist item against the case facts. The system extracts facts $F$ from $q$ and compares them against the requirements $R$ derived by Agent 1. Each item $r$ is marked as either satisfied (requirement clearly met by facts $F$, with the relevant passage quoted), or unaddressed (no direct mention or incomplete information). For items marked unaddressed, the agent assesses relevance: critical gaps are those required by statute or case law or that are outcome-determinative, while items not relevant are considerations that do not apply to the specific case. The agent applies specific legally and logically grounded verification principles: degree and severity require frequency, extent, and intent; comparative requirements need actual numbers; document format requirements need explicit description; contract mentions do not equal violations without explicit statement; and conflicting accounts between claimant and employer reflect critical gaps. 

\underline{{Agent 3 (Supervisory Review)}} serves as the final authority, independently reviewing the requirements generated by Agent 1 and the previous assessment of Agent 2. This agent verifies satisfied items against legal requirements and overturns to unaddressed if legally insufficient. For unaddressed items, the agent independently reassesses whether facts could be reasonably inferred from stated facts and whether gaps are truly outcome-determinative. 

When Agent 3 recommends proceeding, no critical gaps remain, and the system produces a determination grounded in the statutory provisions and case law retrieved in the first component. When Agent 3 recommends abstention, critical gaps exist, and the system outputs ``inconclusive'' along with identification of which specific requirements lack adequate support, and what additional information is needed. This explicit gap identification is critical for fact-finding processes conducted by UI adjudicators.

\subsection{Design Principles}

Our framework enforces specific principles that prevent both presumptuousness and excessive caution. These principles are also incorporated into the enhanced prompting configurations used when evaluating other AI systems. Stated facts that are not contradicted by other stated facts are treated as established; the system does not demand proof of assertions made in the query. However, unstated facts remain unknown---the system does not infer nonexistence from silence or assume missing information favors either party. When claimant and employer accounts directly conflict on outcome-determinative facts, the conflict is flagged for resolution through fact-finding rather than assumed away. Statutory thresholds using comparative language (greater of, exceeds, longer than) are assessed based on the specific values or circumstances provided in the query. When statutes require consideration of specific factors, absence of those factors in the query constitutes a gap preventing determination. Relevant case law principles identified in adjudication guides must be satisfied by the stated facts for determination to proceed. Explicit requirement enumeration transforms the implicit question ``is there enough information?'' into an explicit reasoning process over requirements $R$ and facts $F$.

The multi-agent architecture creates a dependency chain where each component receives outputs from all previous components as explicit input, ensuring that the determination cannot execute until multi-agent assessment completes. Rather than using fixed templates, SPEC generates prompts dynamically based on each query. Following a system prompt, an initial planner LLM analyzes the query and produces tailored instructions that encode awareness of position in the pipeline, explicit reference to outputs from prior components, and analysis strategies specific to the statutory provisions at issue.  

A single-pass approach would require the model to simultaneously determine what facts are needed, whether those facts are present, and what conclusion follows. SPEC decomposes these into distinct components: requirement extraction, fact verification, and supervisory review, each with explicit interfaces between stages.

\section{Experiments}
\label{sec:experiments}

\subsection{Dataset}

We construct a dataset of 250 questions based on Colorado UI law through careful curation grounded in detailed understanding of the governing statutes, regulations, and procedures. As described earlier, our collaboration with CDLE enables us to construct realistic and representative questions that reflect the actual challenges faced in UI adjudication. Questions are designed to test both the ability to reach correct determinations when sufficient information exists and the ability to recognize when critical facts are missing.
The dataset includes both conceptual questions about legal provisions and fact-specific cases requiring eligibility determinations. Questions address scenarios involving voluntary quit, discharge for misconduct, availability for work, and other eligibility criteria under Colorado law. More than half of the cases (56\%) are inconclusive, where critical facts necessary for a determination are intentionally withheld. These cases test whether systems appropriately defer judgment when information is insufficient. The remaining cases provide complete information necessary for accurate eligibility determinations. 


\subsection{Models and Evaluation Metrics}

For our comparisons with existing AI systems, we focus on Claude Sonnet 4.5, GPT 5.2, Gemini 3, and NotebookLM. NotebookLM is Google's document analysis tool designed to answer questions based on uploaded materials, and it has been suggested for legal settings \cite{ouelletteCanAIHold2025a}. All systems receive identical legal materials through RAG for each query. For SPEC, we use Sonnet 4.5 as the underlying model. Additional analyses in this section explore varying levels of information completeness, enhanced prompts, advanced prompting techniques, ablation of SPEC features, and SPEC's performance when running on different underlying models.
Across all analyses, we measure accuracy on two types of cases. Complete cases contain sufficient information for a determination, and we assess whether the system produces the correct outcome, such as eligible/ineligible for benefit eligibility cases or yes/no for other direct questions where sufficient information is provided. Inconclusive cases lack critical information needed for a determination, and we measure whether the system correctly identifies these cases as inconclusive rather than issuing a premature determination. Overall accuracy reflects performance across the entire test set.

\subsection{Baseline Performance and Presumptuousness}

We first evaluate all systems using a standard prompt without specialized instructions to establish baseline performance: 

\vspace{1.5em}

\begin{center}
\parbox{0.94\columnwidth}{%
\rule{\linewidth}{0.5pt}\\[0.3em]
You are an expert on Colorado Unemployment Insurance law. Use the provided document to answer questions accurately and concisely, provide clear determination and reasoning in one paragraph, and answer the exact question asked by the user.\\
\rule{\linewidth}{0.5pt}}
\end{center}

\vspace{1.5em}

\noindent
Table~\ref{tab:baseline_results} presents baseline results across all four systems. Overall accuracy remains low, with GPT 5.2 achieving the highest performance at 0.54 while Sonnet 4.5 reaches only 0.42. More critically, accuracy on inconclusive cases reveals severe presumptuousness across all systems. GPT 5.2 appropriately defers judgment in only 30\% of cases lacking sufficient information, while Sonnet 4.5 defers in just 8\% of such cases. This pattern demonstrates that current systems routinely issue confident determinations even when critical facts necessary for accurate adjudication are absent, reflecting a fundamental failure to recognize the limits of available information.

\begin{table}[h]
  \centering
  \begin{tabular}{lccc}
    \toprule
    Model & All Cases & Complete & Inconclusive \\
    \midrule
    Sonnet 4.5 & 0.42 & 0.84 & 0.08 \\
    Gemini 3 & 0.48 & \textbf{0.90} & 0.15 \\
    GPT 5.2 & 0.54 & 0.85 & 0.30 \\
    NotebookLM & 0.44 & \textbf{0.90} & 0.08 \\
    SPEC & \textbf{0.89} & 0.89 & \textbf{0.89} \\

    \bottomrule
  \end{tabular}
  \caption{Baseline model accuracy on all cases, complete cases, and inconclusive cases.}
  \label{tab:baseline_results}
\end{table}

\subsection{Model Presumptuousness Across Different Levels of Information Completeness}

Figure~\ref{fig:baseline-degradation} illustrates model accuracies across different levels of information completeness. Eligible and ineligible cases contain all facts required for determination. Inconclusive cases are missing one to four critical pieces of information, labeled as Missing-1 through Missing-4, respectively. These comparisons test both whether systems recognize when available information is insufficient for reliable adjudication, and to what degree they can detect gaps. Missing-4 cases with four missing facts should be relatively easier to identify as incomplete, while Missing-1 cases with only one missing fact more closely resemble complete cases and present a harder recognition challenge.

\begin{figure}[h]
\centering
    \includegraphics[width=\linewidth]{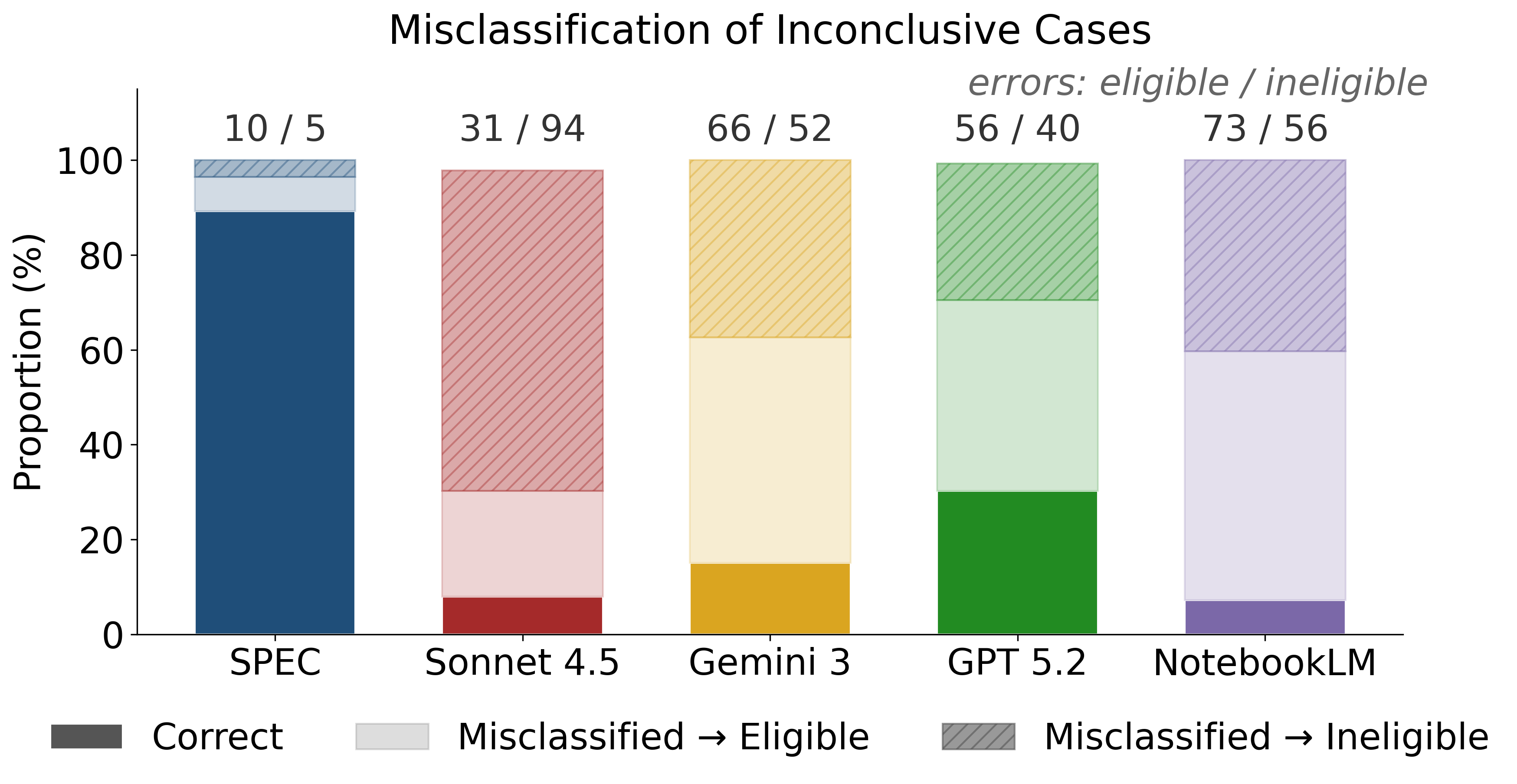}
    \caption{Error analysis of inconclusive cases (n=139). Rates of false ``denial'' or ``approval'' of incomplete cases vary by model.}
\label{fig:error-analysis}
\end{figure}

All models perform reasonably well on complete cases where all information is provided. Sonnet 4.5, Gemini 3, GPT 5.2, and NotebookLM achieve 75-91\% accuracy on eligible cases and 90-96\% on cases where claimants are ineligible. However, performance collapses dramatically on inconclusive cases. Almost all models fall below 35\% accuracy when critical facts are missing, with most performing substantially worse. NotebookLM and Sonnet 4.5 both salvage 7\% accuracy on Missing-1 cases. GPT 5.2 shows the best performance among baseline models on inconclusive cases, maintaining 70\% on Missing-4, but still fails to appropriately defer on the majority of cases lacking sufficient information. This pattern demonstrates that systems routinely issue presumptuous determinations even when critical facts are absent. For a claimant, a false denial means lost income during a period of unemployment; a false acceptance creates an overpayment the agency may later seek to recover. Figure~\ref{fig:error-analysis} shows how failures vary by model. Sonnet 4.5 is especially likely to disqualify outright, issuing unwarranted denials (``Ineligible'') in 68\% of cases with missing information (94/139)---particularly concerning given that our inconclusive cases do not include \textit{any} disqualifying facts. SPEC maintains consistent performance across all categories, achieving 93\% on eligible cases, 90\% on ineligible cases, and 83-100\% across all inconclusive subcategories.

\begin{figure}[h]
    \centering
    \includegraphics[width=\linewidth]{Fig_4_baseline_accuracy_by_category.png}
    \caption{Model accuracy across information completeness levels. Eligible and Ineligible cases contain all necessary facts. Missing-1 through Missing-4 have one to four critical facts missing. Models perform well on complete cases but fail dramatically on inconclusive cases. SPEC maintains high accuracy across all categories. Error bars show 95\% bootstrap confidence intervals.}
    \label{fig:baseline-degradation}
\end{figure}

\subsection{Instruction-Enhanced Models and SPEC}

We evaluate enhanced configurations where models receive explicit adjudication instructions. The most common issues involve the ``able and available'' requirement and wage threshold verification: while verifying availability and confirming claimants earned sufficient wages during their base period are standard practices in adjudication, these are applied across cases and are not the focus of this work. The enhanced prompt (summary in the next paragraph; full version in Appendix) addresses this by assuming claimants are able, available, and meet wage requirements unless explicitly stated otherwise, allowing models to focus on the substantive legal questions that actually determine eligibility. Additional instructions require deferral when critical facts are absent, verification of deadlines and thresholds, systematic statutory checking, and case law application. 

\vspace{1.5em}
\begin{center}
\parbox{0.94\columnwidth}{%
\rule{\linewidth}{0.5pt}\\[0.3em]
Assume claimant able and available for work unless explicitly stated otherwise. When critical information is genuinely missing, output Inconclusive and specify what additional facts are needed; do not assume unstated facts. Verify all time-based requirements including filing deadlines and waiting periods. Check quantitative thresholds such as severity or frequency of misconduct. Identify each statutory requirement and verify satisfaction. Apply relevant case law from the adjudication guide. Provide a clear determination with statutory citations.\\
\rule{\linewidth}{0.5pt}}
\end{center}

\vspace{1.5em}

\begin{figure}[h]

    \centering

    \includegraphics[width=\linewidth]{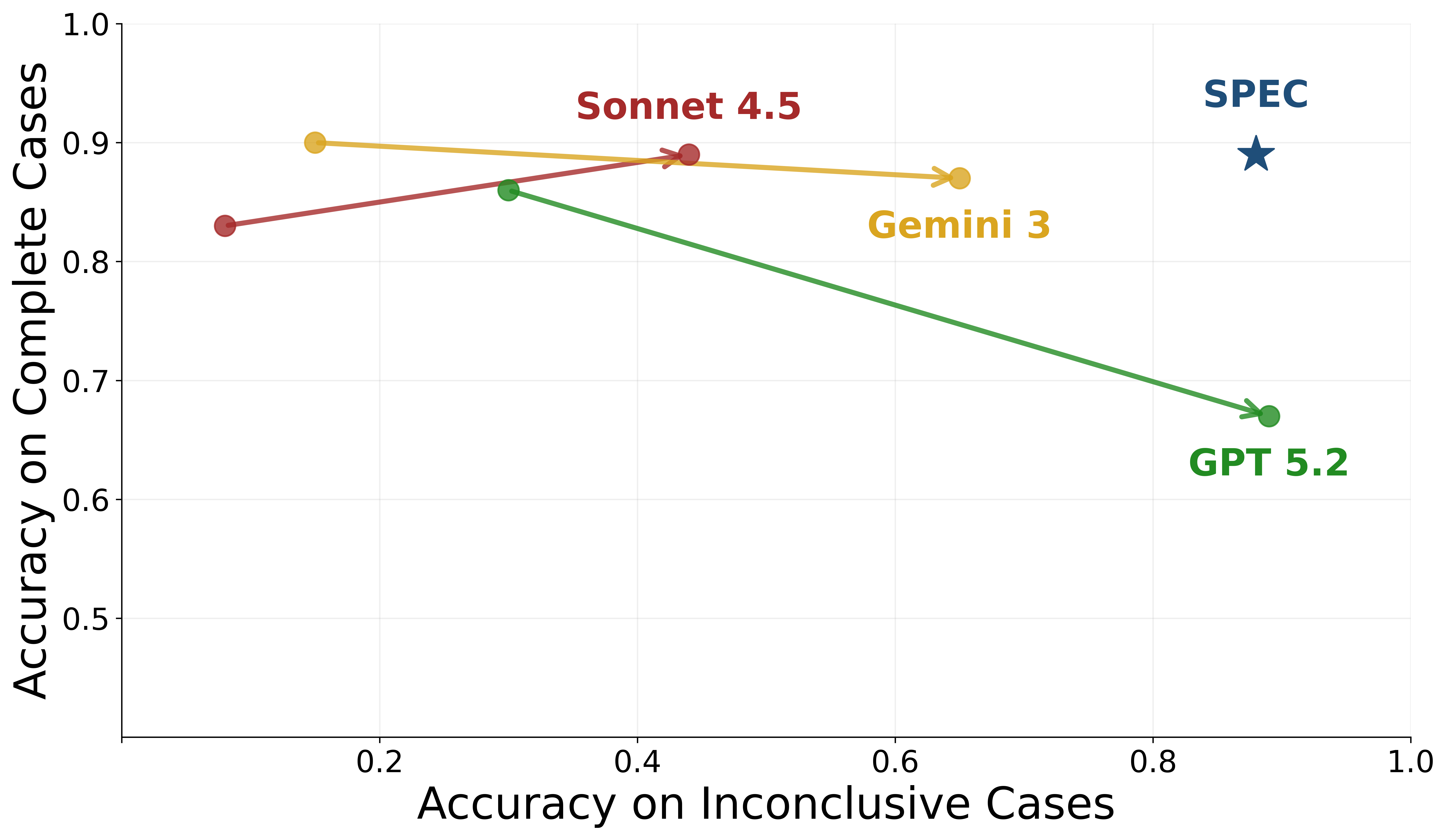}

    \caption{Performance trajectories from baseline to enhanced prompting. For Gemini 3 and GPT 5.2, enhanced prompting improves inconclusive case accuracy but reduces complete case performance (the determination--deferral tradeoff). SPEC achieves a near-Pareto improvement across all performances, maintaining high accuracy on both complete cases (89\%) and inconclusive cases (89\%).}

    \label{fig:base-enhanced}

\end{figure}

\noindent
Table~\ref{tab:enhanced_results} presents performance for enhanced configurations. Enhanced instructions improve overall accuracy across all models. GPT 5.2 Enhanced achieves the highest accuracy on inconclusive cases at 0.90. SPEC achieves 0.89 overall accuracy with strong performance on both complete cases (0.89) and inconclusive cases (0.89). Figure~\ref{fig:base-enhanced} visualizes how enhanced instructions shift model behavior from baseline to enhanced configurations. The gradient trajectories exemplify an underlying determination--deferral tradeoff: models that improve on inconclusive cases (like GPT 5.2, moving from 0.30 to 0.90) tend to become more conservative on complete cases (dropping from 0.85 to 0.68). This pattern suggests that instruction-based approaches to gap recognition face a tension between identifying missing information and making determinations when information is sufficient. While Sonnet 4.5 appears to partially escape this tradeoff, it still fundamentally trails behind in overall accurate deferral, whereas SPEC achieves a near-Pareto improvement across the best of all other model performances.

\begin{table}[h]
  \centering
  \begin{tabular}{lccc}
    \toprule
    Model & All Cases & Complete & Inconclusive \\
    \midrule
    Sonnet 4.5 Enhanced & 0.64 & \textbf{0.90} & 0.44 \\
    Gemini 3 Enhanced & 0.75 & 0.88 & 0.65 \\
    GPT 5.2 Enhanced & 0.80 & 0.68 & \textbf{0.90} \\
    SPEC & \textbf{0.89} & 0.89 & 0.89 \\
    \bottomrule
  \end{tabular}
  \caption{Enhanced model accuracy on all cases, complete cases, and inconclusive cases.}
  \label{tab:enhanced_results}
\end{table}

\subsection{Advanced Prompting Methods}

We evaluate five established prompting techniques using GPT 5.2 to assess their effectiveness at legal adjudication and information gap recognition. All methods receive identical statutory context through RAG and use the enhanced prompt template. 
Table~\ref{tab:prompting-methods} shows that advanced prompting methods achieve modest overall accuracy (0.61-0.71) with similar performance patterns to standard prompting: reasonable accuracy on complete cases (0.76-0.87) but poor deferral rates on inconclusive cases (0.49-0.64). These results show that existing multi-step and multi-agent prompting approaches provide marginal improvements but do not fundamentally address the challenge of recognizing information gaps in legal adjudication.

\begin{table}[h]
  \centering
  \begin{tabular}{lccc}
    \toprule
    Method & All Cases & Complete & Inconclusive \\
    \midrule
    Chain-of-Thought & 0.70 & \textbf{0.87} & 0.57 \\
    Tree-of-Thoughts & \textbf{0.71} & 0.81 & \textbf{0.64} \\
    Self-Consistency & 0.70 & 0.85 & 0.58 \\
    Reflexion & 0.68 & 0.84 & 0.56 \\
    Multi-Agent Debate & 0.61 & 0.76 & 0.49 \\
    \bottomrule
  \end{tabular}
  \caption{Advanced prompting methods accuracy on all cases, complete cases, and inconclusive cases.}
  \label{tab:prompting-methods}
\end{table}

\subsection{Advanced Prompting Methods with SPEC}

We evaluate whether explicitly prompting advanced methods to identify information gaps improves their performance. These configurations incorporate SPEC's core principle of systematic gap identification through modified prompts, though they do not use SPEC's full multi-agent architecture. 


Table~\ref{tab:prompting-methods-GAP} presents results for these gap-aware configurations. Overall accuracy ranges from 0.71 to 0.81, with CoT + SPEC achieving the best performance. Inconclusive case accuracy improves across all methods, reaching 0.84-0.96 compared to 0.49-0.64 in the baseline versions. However, complete case accuracy drops to 0.50-0.74, indicating these configurations become overly conservative in general---in other words, vulnerable to the same determination--deferral tradeoff. 

\begin{table}[h]
  \centering
  \setlength{\tabcolsep}{3pt}
  \begin{tabular}{lccc}
    \toprule
    Method & All Cases & Complete & Inconclusive \\
    \midrule
    Chain-of-Thought + SPEC & \textbf{0.81} & \textbf{0.74} & 0.87 \\
    Tree-of-Thoughts + SPEC & 0.77 & 0.53 & \textbf{0.96} \\
    Self-Consistency + SPEC & 0.79 & 0.72 & 0.84 \\
    Reflexion + SPEC & 0.73 & 0.50 & 0.91 \\
    Multi-Agent Debate + SPEC & 0.71 & 0.50 & 0.87 \\
    \bottomrule
  \end{tabular}
  \caption{Advanced prompting methods with explicit gap detection instructions.}
  \label{tab:prompting-methods-GAP}
\end{table}

\subsection{Ablation Studies}
\label{sec:Ablations}
We isolate the contributions of SPEC's components through systematic evaluation. Table~\ref{tab:ablation} presents results across four configurations. \textit{2-Agent (No Extractor)} merges the extraction and verification functions into one agent while retaining the independent supervisor. \textit{2-Agent (No Supervisor)} merges verification and supervisory review while retaining a separate extraction agent. \textit{Single Agent} collapses all three functions into one agent performing extraction, verification, and review in a single pass. \textit{Static Prompting} replaces dynamic prompt generation with fixed templates: whereas the full SPEC pipeline generates stage-specific prompts adapted to each question, Static Prompting uses identical prompts regardless of the legal issue presented.

\begin{table}[h]
  \centering
  \begin{tabular}{lccc}
    \toprule
    Configuration & All Cases & Complete & Inconclusive \\
    \midrule
    2-Agent (No Extractor) & 0.78 & 0.85 & 0.73 \\
    2-Agent (No Supervisor) & 0.77 & 0.85 & 0.70 \\
    Single Agent & 0.78 & 0.83 & 0.75 \\
    Static Prompting & \textbf{0.84} & \textbf{0.86} & \textbf{0.83} \\
    \bottomrule
  \end{tabular}
  \caption{Ablation results showing accuracy on all cases, complete cases, and inconclusive cases.}
  \label{tab:ablation}
\end{table}

Ablation across multiple configurations reveals which components contribute most to performance. The 2-Agent (No Supervisor) configuration shows the largest performance drop, with inconclusive case accuracy falling to 70\% compared to 89\% for the full pipeline. When fact verification and supervisory review are combined, the system loses the independent check that catches errors in requirement satisfaction. The 2-Agent (No Extractor) and Single Agent configurations show similar performance, maintaining 78\% overall accuracy with 73-75\% on inconclusive cases. Static Prompting achieves performance closest to the full pipeline at 84\% overall with 83\% on inconclusive cases. These results suggest that maintaining the three-agent separation is the most important architectural feature, with the supervisory agent providing the most critical contribution to gap detection. 


\subsection{SPEC with Different Underlying Models}

\begin{figure}[h]
    \centering
    \includegraphics[width=\linewidth]{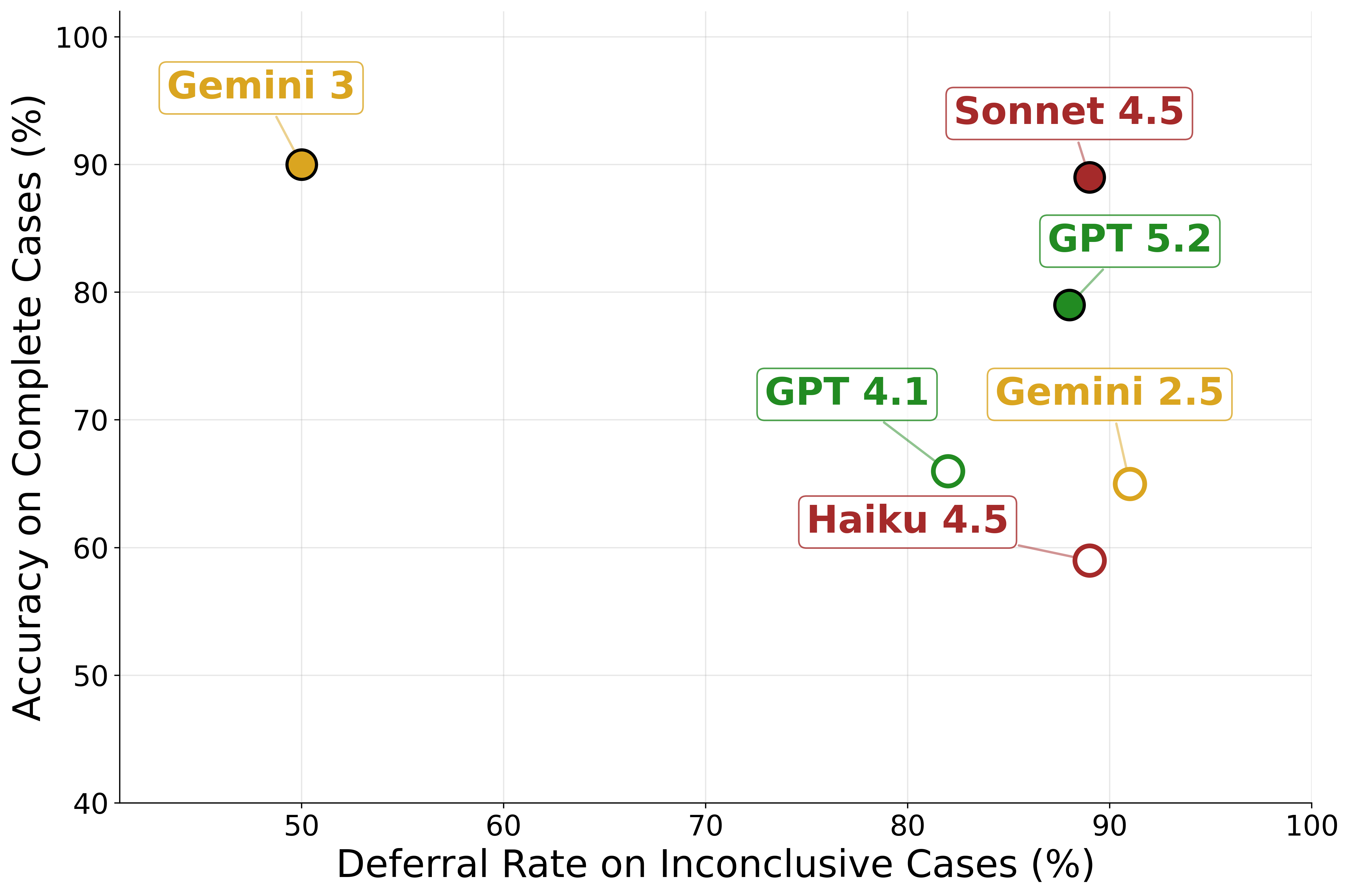}
    \caption{SPEC performance when running on different underlying foundation models. Models show varying tradeoffs between complete case and inconclusive case accuracy. Color indicates provider family.}
    \label{fig:SPEC_models}
\end{figure}

We evaluate SPEC when running on different underlying models to understand how model capability affects performance on both deferral and determination accuracy. This analysis addresses practical deployment considerations where agencies must balance performance against resource constraints. We implement SPEC with six models: Claude Sonnet 4.5, Claude Haiku 4.5, Gemini 3, Gemini 2.5, GPT 5.2, and GPT 4.1. Each configuration receives identical statutory context and case queries. Figure~\ref{fig:SPEC_models} presents results across model tiers. Sonnet 4.5 demonstrates balanced performance across both complete and inconclusive cases. Gemini 3 shows a tendency toward issuing determinations, performing strongly on complete cases but less reliably identifying information gaps. GPT 5.2 exhibits more conservative behavior, readily deferring on inconclusive cases while being more cautious on complete cases. 


\section{Limitations}
\label{sec:limitations}

We evaluate Colorado UI eligibility, a domain chosen for its statutory complexity, real-world deployment pressures, and the rare access our collaboration affords to official training materials. While UI adjudication shares structural features with other administrative settings---vague standards, incomplete evidence, high stakes for individuals---extending this work to other jurisdictions and legal domains would test the boundaries of what we've found and reveal challenges specific to other domains.

Our inconclusive cases were constructed by systematically withholding facts from complete scenarios. Real-world incompleteness is messier: ambiguous rather than absent information, conflicting accounts, and facts whose legal relevance is itself contested. Testing on naturally occurring cases---drawn from actual adjudication records---would show whether SPEC's structured gap detection handles ambiguous uncertainty as well as absence.

SPEC's multi-agent architecture increases computational cost relative to single-pass prompting, a tradeoff worth examining as these tools move toward production. We did not test interactive use, where adjudicators query the system iteratively and refine their understanding; such settings may be where explicit gap identification proves most valuable. Finally, the adjudication guides central to our approach cannot be publicly released, limiting reproducibility. We release our benchmark dataset of carefully-constructed questions and our evaluation framework to support further work.

\section{Conclusion}
\label{sec:conclusion}

Current AI systems can be presumptuous, issuing confident determinations when critical information is missing. Our evaluation of four leading platforms on UI eligibility reveals baseline accuracy averaging 15\% on cases that should be deemed inconclusive. In other words, the AI systems most readily available to UI adjudicators today are extremely likely to provide definitive answers when the appropriate response is, in fact, ``more facts needed.'' 

Enhanced prompting and other advanced prompting methods improve gap recognition relative to base models, but often fall victim to the determination--deferral tradeoff: improvements in the ability to defer appropriately on incomplete cases come at the direct expense of increased conservatism (and hence lower accuracy) on complete cases. 

We introduce SPEC, a framework that escapes this tradeoff through multi-agent analysis requiring explicit identification, enumeration, and verification of information gaps before permitting determinations. SPEC achieves 89\% accuracy on both complete and inconclusive cases---a near-Pareto improvement over the determination and deferral accuracies of all other methods. 

Our findings have direct implications for AI deployment in legal adjudication. Presumptuousness has persisted in frontier models, perhaps abetted by the failure of popular AI benchmarks to recognize and reflect the importance of epistemic humility in many challenging real-world domains. Many existing legal AI benchmarks themselves are designed with 100\% of tasks resolving to a determinable outcome. While we instead designed our evaluation to be a roughly 50/50 split between complete and incomplete cases, a recent study of UI found that real adjudicators chose to seek additional fact-finding in 87\% of claims \cite{mageshEvaluatingGenerativeAI2026}, suggesting that the accuracy gains of SPEC over other methods in a real-world adjudication setting are likely even greater than what we present in our study. 

SPEC is not architecturally complex; neither is RAG nor Chain-of-Thought. But it assembles targeted design principles that prove effective in practice, and we offer it as a practically useful tool for high-stakes legal reasoning. Our work also informs the longstanding debate around the availability of agency guidance materials.  Such guidance materials may be important to enable responsible and open AI development and research in legal settings, just as open data has fostered public innovation. We hope that our benchmark and SPEC inspire others to address presumptuousness and other challenges with clear real-world stakes.

\section*{Generative AI assistance}
We used ChatGPT and Claude for assistance with editing and for supporting scripting tasks. 

\section*{Supplementary Information}
Code and data are available at \url{https://github.com/reglab/spec}.


\bibliographystyle{ACM-Reference-Format}
\bibliography{references}

\end{document}